\documentclass[
]{ceurart}

\usepackage{listings}

\begin{document}

\copyrightyear{2024}
\copyrightclause{Copyright for this paper by its authors.
  Use permitted under Creative Commons License Attribution 4.0
  International (CC BY 4.0).}

\conference{CREAI 2024: International Workshop on Artificial Intelligence and Creativity,
  October 19--24, 2024, Santiago de Compostela, Spain}

\title{THInC: A Theory-Driven Framework for Computational Humor Detection}

\author[1]{Victor {De Marez}}[%
email=victor.demarez@uantwerpen.be
]
\fnmark[1]
\cormark[1]
\address[1]{Centre for Computational Linguistics and Psycholinguistics, University of Antwerp, Antwerp, Belgium}
\address[2]{Department of Computer Science; Leuven.AI, KU Leuven, Leuven, Belgium}
\address[3]{Université de Montréal \& Mila, Montreal, Canada}
\author[2]{Thomas Winters}[]
\author[3]{Ayla {Rigouts Terryn}}[]

\cortext[1]{Corresponding author.}
\fntext[1]{Work partially fulfilled at Department of Computer Science; Leuven.AI, KU Leuven, Leuven, Belgium, and at Centre for Computational Linguistics; Leuven.AI, KU Leuven, Leuven, Belgium.}
\begin{abstract}
Humor is a fundamental aspect of human communication and cognition, as it plays a crucial role in social engagement. Although theories about humor have evolved over centuries, there is still no agreement on a single, comprehensive humor theory. Likewise, computationally recognizing humor remains a significant challenge despite recent advances in large language models. Moreover, most computational approaches to detecting humor are not based on existing humor theories.
This paper contributes to bridging this long-standing gap between humor theory research and computational humor detection by creating an interpretable framework for humor classification, grounded in multiple humor theories, called \textbf{THInC} (Theory-driven Humor Interpretation and Classification).
THInC ensembles interpretable GA$^2$M classifiers, each representing a different humor theory.
We engineered a transparent flow to actively create proxy features that quantitatively reflect different aspects of theories.
An implementation of this framework achieves an F1 score of 0.85. The associative interpretability of the framework enables analysis of proxy efficacy, alignment of joke features with theories, and identification of globally contributing features. This paper marks a pioneering effort in creating a humor detection framework that is informed by diverse humor theories and offers a foundation for future advancements in theory-driven humor classification. It also serves as a first step in automatically comparing humor theories in a quantitative manner.
\end{abstract}


\maketitle

\section{Introduction}
Humor is integral to daily life and human interactions, influencing trust and social bonds~\cite{nijholt2003humor}. The importance of humor for human relationships sparked interest in the domain of human-machine interaction, as the ability to handle humor can make systems appear more friendly and competent~\cite{morkes1998humor}. This highlights the need for computational humor models, which have promising applications in areas like edutainment, service robots, chatbots, humor translations, and recommendation systems~\cite{nijholt2017humor,raskin2002quo,winters2021computers}. In order to realize this, a system capable of detecting or generating humor is required.

There are two distinct influences in the field of computational humor research: recent AI methods making their way into humor research, and humor researchers building computational humor theories and humor systems. The main difference lies in their approach towards computational humor: %
whereas the former rarely consider theoretical humor theories, the latter have humor theories as the foundation of their systems~\cite{amin-burghardt-2020-survey}. 

A humor theory is a \emph{theory} of what is funny and what is not. Humor theories are not theories in the strict sense, however. They are too vague, too broad, or incomplete~\cite{raskin2017humor}. Their lack of specificity makes them difficult to falsify~\cite{MARTIN201833}. There is no universally accepted humor theory that encompasses all genres of humor, even though theories of humor have been around since the Classical Antiquity~\cite{larkin2017overview}%
. Multiple theories were developed over the ages, and new ones still emerge~\cite[\textit{inter alia}]{suls1972two,Raskin1984,benignviolationtheory,toplyn2023witscript}. 

Despite the inherent problems, humor researchers who investigate computational humor often use these humor theories as the foundation for their work. One of their research areas is to create \emph{computational humor theories}, i.e., humor theories that are supposed to be computable. These theories are based on, or at least inspired by %
fundamental humor theories, which makes them interesting as foundations for theory-guided humor systems. However, research in this area is limited and none of these theories can actually serve as a solid, unambiguous basis for an implementation in a humor system. Some suffer from the same limitations as humor theories, as they are too vague or have limited applicability, whereas others outsource the core work needed for humor recognition, or they are simply not computable~\cite{ritchie1999developing,toplyn2023witscript,toplyn2023witscript3,raskin2017script}. Another research area is to develop humor systems that do try to integrate humor theory. This is, however, usually done in a way that makes it hard to see the parallels between the theory and the implementation ~\cite{Ritchie_2009}.

The majority of humor systems are created independently by the NLP research community and do not leverage established humor theories that have been refined over time~\cite{winters2021computers,amin-burghardt-2020-survey}. This research focuses more on performance and computational feasibility, rather than on acknowledging and integrating the nuances of humor and humor theories~\cite{Ritchie_2009}.%

The discussion above highlights a research gap characterized by three possible scenarios: performance that lacks a theoretical foundation and thus fails to build on the current understanding of humor, theory that is not readily translatable into a practical, computational system, or a computational system that, despite having a theoretical underpinning, lacks clear connections between the theory and its implementation. This paper aims to contribute to bridging this gap by posing the following two research questions:
\begin{itemize}
    \item \textbf{RQ1}: Is it possible to construct a framework that learns to detect humor by leveraging %
    humor theories to a maximal extent, while still maintaining strong performance?
    \item \textbf{RQ2}: Can we associatively trace what is learned by this framework back to their underlying theoretical concepts of humor? 
\end{itemize}

We highlight the machine-learned associative nature of the backtracing in \textbf{RQ2}, as opposed to causal or ontological humor detection approaches. To date, there has been only one attempt to employ ontologies in humor detection, which theoretically enables a deeper computational understanding of humor, but proves intractable in practice~\cite{raskin2017script}. Consequently, approaches like ours, which are non-ontological, avoid these methods, ensuring computational feasibility but potentially sacrificing the depth of understanding that ontologies could provide, in favor of machine-learned associations.

The remainder of this paper is structured as follows. Section \ref{sec:background} offers some background on humor theories and the machine learning model used. Section \ref{sec:relatedwork} provides an overview of existing humor detection systems that incorporate elements of humor theory. In Section \ref{sec:methodology}, we provide a detailed description of our framework's classifiers, the interpretability mechanism, and the flow of engineering and calculating proxy features that capture part of humor theories. In Section \ref{sec:experiments}, we implement a concrete system with the presented architecture. The implementation is evaluated in Section \ref{sec:results}, answering the research questions. %
Finally, the conclusion in Section \ref{sec:conslusion} is followed by Section \ref{sec:futurework}, which explores potential future modifications of the framework. 

The source code of the implementation of the framework used for evaluation is available online (\url{https://doi.org/10.5281/zenodo.13366981}).

\section{Background}\label{sec:background}

\subsection{Humor Theories}
As mentioned, there is no universally accepted theory of humor.
The three established theories of humor are distilled from research lines over centuries, and therefore lack a unified definition in literature, explaining their inherent ambiguity and vagueness.
An attempt to define them is as follows, based on descriptions by~\citet{larkin2017overview,meyer2000humor,buijzen2004developing}:
\begin{itemize}
    \item The \textbf{superiority theory} suggests that those who see themselves as superior laugh at inferiors and wrongdoers, reinforcing social divisions and maintaining societal order. This laughter boosts the confidence of the one laughing, manifesting in joyfulness and more laughter.
    \item The \textbf{relief theory} suggests that laughter results from the release of built-up psychological tension, transforming into muscle movement. This swift change from intense to reduced tension leads to joy. The tension may stem from excitement, an uneasy state of arousal, or from stress, which heightens arousal.
    \item According to the \textbf{incongruity theory}, people laugh when there's a violation of an expected pattern, an unexpected twist or incongruity, or a surprise. This unexpected turn must be non-threatening yet sufficiently abnormal to be noticed.
\end{itemize}

Under the influence of several twentieth-century thinkers, the incongruity theory is extended to the incongruity resolution (IR) theory, which divides the humor process into two stages: the introduction of an incongruity, and its resolution, by applying a different cognitive rule~\cite{larkin2017overview}.
Another more computational but widely respected extension of the incongruity resolution is the \textbf{surprise disambiguation (SD) model}~\cite{ritchie1999developing}.
It is important to note that the SD model is more concrete than the IR theory, therefore excluding manifestations that were actually explainable by the IR theory with a stretch.
Conversely, the SD model, with its distinct perspective, brings forth straightforward manifestations that might appear implausible when viewed through the IR theoretical framework.
These four humor theories are equally respected, with their own paradigms on what constitutes humor~\cite{sen2012humour}.

\subsection{Generalized Additive Model Plus Interactions}
A GA$^2$M model (generalized additive model plus interactions) is a white box machine learning model.
Such a model has the form
\begin{equation}
    g(E[y])=\beta_0+\sum_if_i(x_i)+\sum_{i\neq j}f_{ij}(x_i,x_j),
\end{equation}%
where $g$ is the logistic link function, $x_i$ is the $i$th feature in the feature space, and $f_i$ is the corresponding feature function~\cite{lou2013accurate}. The feature functions are shallow bagged trees trained with
gradient boosting~\cite{lou2012intelligible}. GA$^2$M have high accuracy compared to regular GAM models due to the addition of two-dimensional interactions. %
\section{Related Work}\label{sec:relatedwork}
\begin{figure*}[ht]
	\centering
	\includegraphics[width=1.01\textwidth]{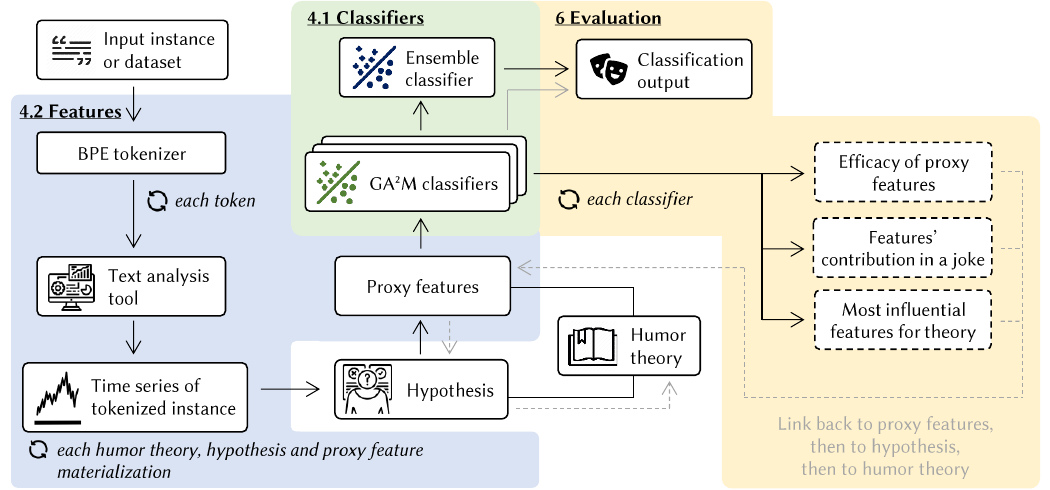}
	\caption{Graphical representation of the architecture of the THInC framework. The architecture consists of three different modules: feature generation, GA$^2$M classifiers and classification results, and interpreting the learned feature application in a classifier on its efficacy of capturing its humor theory. The \includegraphics[height=7pt]{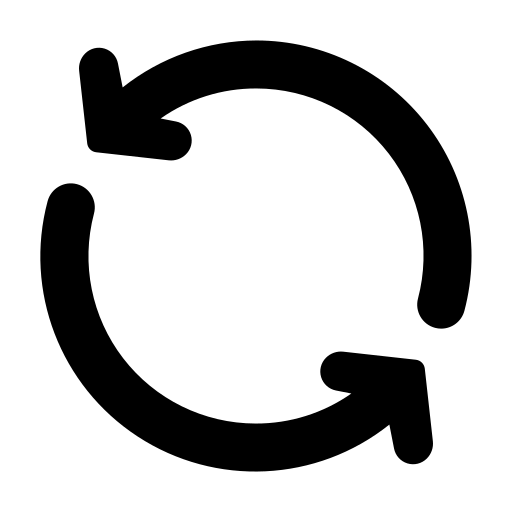} symbol indicates the possibility of repeating (consecutive) flows over the adjacent quantification.}
	\label{fig:architecture}
\end{figure*}
There have been many approaches to perform automatic humor detection.
Earlier approaches aimed to distinguish jokes from unrelated types of texts such as news and proverbs by relying on simple models using word-based features, sometimes inspired by humor theory~\cite{mihalcea2005svm,taylor2004computationally}.
Seminal approaches by humor researchers that integrate a single humor theory include the detection of wordplay or recognition of the punchline in one-liners~\cite{taylor2004computationally,mihalcea2010computational}.
More modern approaches generally use large language models such as BERT to distinguish jokes from non-jokes~\cite{annamoradnejad2020colbert,winters2020robberthumor}, but they typically do not use humor theory-informed features or architectural design.
As such language models already contain a lot of linguistic knowledge, they tend to exhibit strong performance even without explicitly integrating theoretical knowledge. Moreover, they generally lack the ability to incorporate such symbolic knowledge.

Current humor detection systems either focus on limited types of humor, only encompass a limited view of humor, or use features in their implementation that do not fully align with a humor theory, thus contaminating the results if one were to try to measure the importance of humor theories with such a system.

\section{Framework Architecture}\label{sec:methodology}
Figure \ref{fig:architecture} provides an overview of the THInC framework's architecture. The humor detection approach of the framework differentiates jokes from non-jokes through a faithful interpretable feature-based approach, rather than a black-box language modeling approach, giving us insights into the humor process.

\subsection{Classifiers}\label{sec:classifiers}
Our framework approaches humor detection as a binary classification task.
Each humor theory is represented by a GA$^2$M classifier and trained on a distinct feature set, making the framework adaptable to various humor theories.
The outcomes of each classifier are then combined to a final prediction through the application of an ensemble model.
\subsubsection{Backtracing to Humor Theories}
GA$^2$M models are particularly suitable for our framework due to their interpretability, stemming from the modularity of the additive model. The additivity ensures that the marginal contribution of each function $f_i$ and $f_{ij}$ can be understood. Single features can be understood due to the \emph{function shaping} nature of GAM models, where each feature value $k$ of feature $i$ has a corresponding function value $f_i(k)$, which is the logit contribution to the prediction of an instance. Hence, a feature function can be easily plotted, providing interpretability on the feature level ~\cite{lou2013accurate}. %
Pairwise interactions can be understood analogously through a heat map~\cite{lou2013accurate}. Interpretability on the global and local levels is possible due to derivations of the feature level interpretability.

An interpretability analysis helps to evaluate and refine the link between the features and the humor theory. A trade-off inherent to basing humor detection on these ambiguous and vague theories is therefore the maximal possible theoretical depth of the evaluation.
The evaluation and interpretable power are further discussed in Section \ref{sec:results} on a concrete implementation of the architecture.

\subsection{Features}\label{sec:features}
Humor theories are inherently ambiguous, and they can manifest in diverse ways, complicating leveraging them for detection. The degree to which a manifestation is pertinent to a particular joke often depends on the interpreter's willingness to stretch the theory to apply. While this flexibility allows for a broad range of interpretations, it also carries the risk of generating false positives. Regardless of the underlying willingness required for these theories to perform optimally, it is essential for any computational implementation to define these manifestations clearly. For a humor detector to be effectively rooted in these theories, it must encompass their explicit manifestations as comprehensively as possible. Manifestations that prove to be inaccurate are filtered out during the training phase, as they would be assigned a negligible feature function value. %

In our framework, a manifestation of a theory takes the form of a feature, which we term a \emph{proxy feature} or \emph{proxy}. 
The process of engineering and computing features that embody a humor theory in a specific manner while ensuring they are computable is an integrated workflow. A worked-out example of the workflow below in our implementation of the framework can be found in Section \ref{subsec:features}.
\begin{enumerate}
\item Identify computational tools or mechanisms that analyze text at a word level or lower.
\item Tokenize the instance with any tokenizer. The tokenization step splits text to account for the temporal structure inherent to humor theories.
\item Depending on what is sensible for the selected computational tool, create a time series with a value for each token of the instance by doing exactly one of the following:
\begin{itemize}
\item \textbf{Token-based}: Calculate a value for each token in the instance by passing just that token to the computational tool.
\item \textbf{Subsequence-based}: Calculate a value for each token in the instance based on each prefix subsequence up to that token using the computational tool. This means that the first value is calculated on the first token, the second value uses the first two tokens, etc. For example: execute the subsequence-based approach with the anger detection model, where each value is the probability of that prefix subsequence being angry. This type of left-contextual features mimics how humans hear jokes and helps model the perception of the joke over time.
\end{itemize}
This step captures the temporal nature of humor theories, which inherently unfold over time with clear beginnings and endpoints, similar to the way a joke is delivered.
\item Formulate straightforward hypotheses linking the characteristics in the time series to qualitative aspects of a humor theory.
\item From the time series, extract numerical proxy features that quantitatively represent these hypotheses.
\end{enumerate}
Iterate this process across various computational tools, proxy features, hypotheses, and theories to create a comprehensive set of proxies that might capture different elements of humor theories.

\section{Implementation}\label{sec:experiments}
To illustrate the practical applications beyond theoretical concepts, we apply the THInC framework in a concrete environment with computational tools available today. The application is an implementation of the framework to demonstrate its working and interpretable power, and is one of many possibilities. The implementation involves using four widely recognized theories of humor, incorporating 155 features, and ensembling GA$^2$M classifiers with a soft voting classifier.

The architecture is implemented systematically in three main steps. First, we identify potential features along with their associated hypotheses and implement the calculations to derive these features, following the workflow outlined in Section \ref{sec:features}. Next, we execute the feature implementations on the dataset, and we use the computed features to train the individual classifiers as well as the ensemble classifier. Finally, we leverage the test portion of the dataset to evaluate and interpret the trained classifiers of our implementation.

\subsection{Data}\label{subsec:data}
We use the dataset of jokes and nonjokes from \textit{SemEval 2021 Task 7: HaHackathon, Detecting and Rating Humor and Offense}~\cite{meaney-etal-2021-semeval}. %
The training set has 8000 instances with a joke-non-joke ratio of 1:0.62. The training set has 1000 instances with the same class imbalance. The validation set also has 1000 instances but with an imbalance ratio of 0.58. Labels are given by 20 annotators. The origin of the data is 80\% from Twitter and 20\% from a dataset of short jokes.

\subsection{Features}\label{subsec:features}
Strictly following the workflow in Section \ref{sec:features}, we engineer and implement 155 features%
, of which two are exemplified in detail below. All features can be found in Appendix~\ref{app:features}. In summary, for this calculation, we use 10 computational text analysis tools.
Seven are large language models equipped with a classification head. They process input via a subsequence-based approach and detect polarity, emotions, offense, subjectivity, hate, stance, and adult language.
Another tool is the LLaMA-2 large language model with a language modeling head ~\cite{touvron2023llama}. The remaining two are custom implementations, detecting ambiguity and morphosyntactic ambiguity. For these last three tools, we employ a token-based method. The tokenization method of choice in this implementation is Byte-Pair Encoding (BPE) tokenizer~\cite{sennrich2016bpe}.

We form hypotheses about how the 10 resulting time series might embody aspects of various humor theories. %
Features are only implemented for the four established and fundamental humor theories that have stood the test of time: the superiority theory, the relief theory, the incongruity theory, and the incongruity resolution theory. We do not impose our own definition of what constitutes a joke.

To capture these hypotheses as accurately as possible, we use the \texttt{tsfresh} library in Python, which implements feature calculations on time series~\cite{christ2018time}. This allows us to extract a range of numerical proxy features from each time series, ensuring the hypotheses were represented effectively.

An example of two implemented subsequence-based features, following the feature engineering workflow, is as follows:
\begin{enumerate}
\item To detect the probability of joy, optimism, anger, and sadness in a text, we identified the emotion recognition model of TweetNLP~\cite{camacho-collados-etal-2022-tweetnlp}.
\item We tokenize the dataset introduced in Section \ref{subsec:data} with a Byte-Pair Encoding (BPE) tokenizer.
\item We use a subsequence-based calculation technique, feeding an increasingly larger prefix subsequence of BPE tokens to the emotion recognition model, creating four time series of emotions for each dataset instance.
\item We formulate the following two straightforward hypotheses linking time series to a humor theory, amongst others in the implementation:
\begin{enumerate}
\item A manifestation of the incongruity theory is bursts of anger.
\item A manifestation of the relief theory is increasing optimism.
\end{enumerate}
\item Two numerical proxy features calculated with the \texttt{tsfresh} library that represent the above hypotheses are the following:
\begin{enumerate}
\item Bursts of anger are calculated through the maximum change between two consecutive values (\texttt{anger\_max\_change}).
\item Increasing optimism is calculated through the slope of a linear fit of the time series (\texttt{optimism\_linear\_fit\_slope}).
\end{enumerate}
\end{enumerate}
The efficacy of the two example proxy features is assessed in Section \ref{sec:results}.

\subsection{Classifiers}
We use the GA$^2$M implementation provided by the \texttt{interpretML} software library~\cite{nori2019interpretml}. The four GA$^2$M classifiers are trained with default parameters, with two exceptions: we set the number of interactions to the maximum possible and limited the maximum number of bins per feature to 100. The choice to maximize interactions does not compromise interpretability, as pairwise interactions remain visually representable and the features continue to be grounded in humor theory. The restriction on bins is a response to the dataset's limited size, aiming to aggregate more data in each bin. These classifiers were trained using both the training and validation sets. The learning rate is set by default at 0.01, with an early stopping tolerance of 0.0001.

The ensemble classifier is chosen to be a weighted soft voting classifier because it accounts for the uncertainty in each humor theory classifier. The weights are learned with the Nelder–Mead method using the average precision score as the objective function. The prediction is then calculated as
\begin{equation}
    \hat{y} = \arg \max_i \sum^{m}_{j=1} w_j p_{ij},
\end{equation}
where $w_j$ is the weight for the $j$th GA$^2$M classifier and $i\in\{0,1\}$.

\section{Evaluation}\label{sec:results}
In this section, we evaluate our implementation by answering the following experimental questions (EQ):
\begin{itemize}
    \item \textbf{EQ1}: What is the performance of our humor detector implementation?
     \item \textbf{EQ2}: Which humor theory classifier contributes the most to the ensemble classification?
     \item \textbf{EQ3}: How well does a proxy feature capture a humor theory? How can bad proxy features be remedied?
    \item \textbf{EQ4}: To what extent do the proxy features contribute to the detection of a joke?
     \item \textbf{EQ5}: Which proxy features contribute most to capturing a humor theory?
\end{itemize}
The combination of \textbf{EQ3} (proxy feature to humor theory) and either \textbf{EQ4} (local dataset instance to proxy feature) or \textbf{EQ5} (global aggregate of proxy features) yields the maximally achievable interpretable power of the framework to create a direct associative backward link from jokes to humor theories, within the external limits of ambiguous humor theories that can only be interpreted through implicit or, in our case, explicit hypotheses, and of currently available computational tools or mechanisms. 

The focus of the framework, and by extension of the evaluation of this implementation, is on in-domain data, i.e., well-formed sentences that are either jokes or non-jokes. 

\subsection{EQ1: Performance Results}
\begin{table}
    \caption{Results for each of the classifiers in our implementation of the framework outline in Section \ref{sec:experiments}. The F1 score is calculated on the test set of the positive class. The last classifier is a RoBERTa model with a modified head and serves as a black-box benchmark model.}
	\label{tab:resultsf1}
    \begin{tabular}{lrr}
        \toprule
        Classifier & F1 & Weight \\
        \midrule
        Ensemble     & 0.851     & - \\
        \hline
        Incongruity theory     & 0.796     & 1.342  \\
        Surprise disambiguation model  & 0.794     & 0.573 \\
        Superiority theory & 0.786    & 0.442 \\
        Relief theory & 0.818     & 1.591 \\
        \hline
        \textit{Benchmark: RoBERTa + modified head} & \textit{0.943} & - \\
        \bottomrule
    \end{tabular}
\end{table}
The F1 scores and ensemble weights for each classifier, following the training process, parameters, and data described in Section \ref{sec:experiments}, are presented in the first five rows of Table \ref{tab:resultsf1}. The last model in the table acts as a benchmark, demonstrating the best possible classification performance achievable on the dataset by an advanced black-box model. This benchmark model is a RoBERTa model modified with an additional dense layer and multi-sample dropout for the classification of the \emph{[CLS]} token. These modifications mirror those used in the top-performing system of the SemEval-2021 Task 7, as reported by~\citet{song2021deepblueai}. The model was fine-tuned from the \texttt{roberta-large} checkpoint with a 2e-5 learning rate, a batch size of 16, and a weight decay of 0.01, over 10 epochs.

The ensemble model outperforms each individual theory classifier in terms of F1 score (0.851), illustrating the benefits of combining classifiers based on various humor theories.  The improvement may stem from each humor theory explaining some jokes better than others. An ensemble approach captures the strengths of multiple theory classifiers, maximizing prediction power within the limitations of the proxies.

The F1 score of the benchmark model surpasses that of our ensemble. This illustrates an expected trade-off: our system is more interpretable and grounded in theory but cannot leverage the more advanced black box design responsible for the benchmark system's better performance.%

\subsection{EQ2: Ensemble Weights}
In our implementation, the relief theory classifier contributes the most to the prediction of a joke, followed by the incongruity theory classifier, the surprise disambiguation model, and the superiority theory. Notice that the weights presented in Table \ref{tab:resultsf1} are the result of our specific implementations of humor theories and are relative to each other. Thus, while indicative, they should not be interpreted as evidence of the superiority of one humor theory over another. Rather, they indicate that with those weights, the ensemble average precision was found to be optimal in terms of predictive performance within the limits of our chosen proxy features.

\subsection{EQ3: Assessing the Efficacy of Proxy Features}\label{sec:featurelevel}

\begin{figure}
	\centering
	\includegraphics{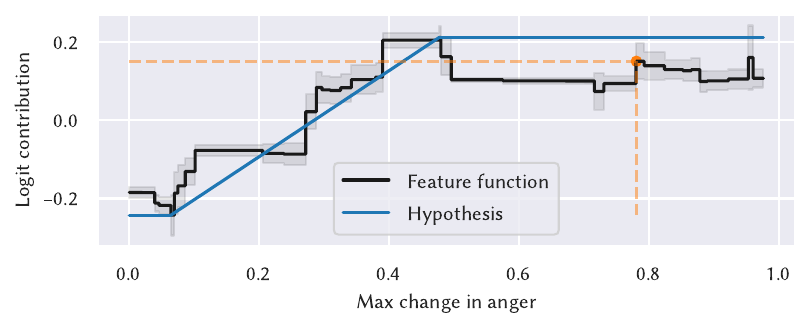}
	\caption{The feature function of `maximal change in anger' in the incongruity theory classifier is shown in black, with the minimal and maximal logit values of the bagged trees in gray, serving as uncertainty intervals. One possible numerical representation of the hypothesis -- that an increase in anger change correlates with a higher likelihood of incongruity -- is presented in blue, across all possible values an instance can have. There is a high correspondence between the actual feature function and the hypothesized feature function. The logit contribution of this feature in test joke 194 (formatted in bold in Figure \ref{fig:localexample_barplot}), corresponding to a specific value for this feature, is marked by an orange dot.}
	\label{fig:featurefunction}
\end{figure}
Interpreting a learned proxy feature allows for assessing its efficacy. %
Figure \ref{fig:featurefunction} illustrates a feature function learned by our incongruity theory classifier for the `maximal change in anger' proxy feature. The hypothesis behind this feature is that any large change in anger correlates with a higher likelihood of incongruity. The blue line depicted in the figure represents one way this hypothesis can be quantitatively expressed.

For every possible quantitative realization of any hypothesis, the exact value of the logit contribution is subordinate to its sign. In this case, small maximal anger changes should get a very negative logit contribution, whereas large maximal anger changes should get a very positive logit contribution. Therefore, the assessment of a proxy feature's effectiveness in representing a humor theory should focus on parts of a feature function with large logit contributions and narrow uncertainty intervals.%

\paragraph{Matching Hypothesis and Reality} With the above consideration, the actual feature function depicted in Figure \ref{fig:featurefunction} closely matches one of the hypothesized feature functions, so it can be reasonably assumed that this proxy feature successfully captures the hypothesized part of the incongruity theory. Notice, however, that this is only the case for a perfectly noise-free proxy feature. That is, however, unattainable, so we choose to overlook this requirement. The assumption can be reinforced by qualitatively validating the hypothesis at large logit contribution points of the proxy feature function through local examples at those points.

\begin{figure}
	\centering
	\includegraphics[]{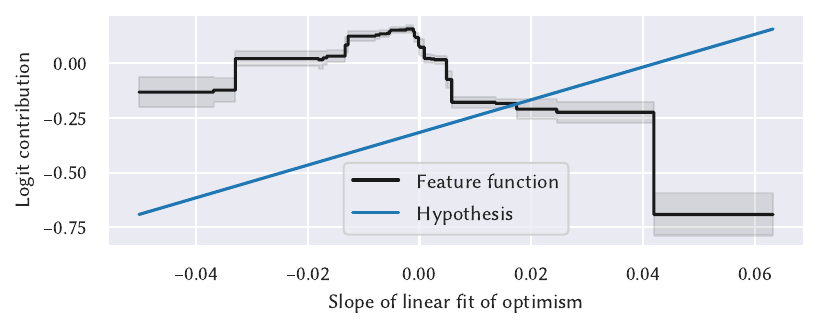}
	\caption{The feature function of `slope of linear fit of optimism' in the relief theory classifier is shown in black, with the minimal and maximal logit values of the bagged trees in gray, serving as uncertainty intervals. One numerical representation of the hypothesis that the slope of linear fit should be positive for the optimism (as a proxy for relief) to increase, is presented in blue across all possible values. There is a low correspondence between the actual feature function and the hypothesized feature function.}
	\label{fig:featurefunction_bad}
\end{figure}

\paragraph{Non-Matching Hypothesis and Reality} An example where the actual feature function does not match with any hypothesized feature function is visualized in Figure \ref{fig:featurefunction_bad} for the `slope of linear fit of optimism' proxy feature in the relief theory classifier. Two independent factors, which may coexist, could be responsible for this phenomenon:
\begin{itemize}
\item \textbf{The hypothesis is wrong.} The interpretation of the theory can be opposite to the engineered hypothesis in that feature. The solution is to find a new hypothesis based on the reality for the relief theory, by looking at local examples. If there is no suitable hypothesis that has a straightforward link with the humor theory, the underlying computational tool should not be used in the classifier.
\item \textbf{The proxy features are noisy.} If the current feature function is inaccurate because the proxy feature fails to measure the intended aspect, more effective proxy features should be used.
\end{itemize}

\subsection{EQ4: Interpreting Features' Contribution in a Joke}\label{sec:locallevel}
Local instances instantiate abstract feature functions at particular values, grounding them in specific semantic meanings that can be validated against humor theories. Some features may have a negative impact due to various reasons: the inherent trade-off between variance and bias, a non-match between the hypothesized and real feature function for the applicable region, or the general reality that not all hypotheses are universally applicable, much like humor theories. However, this is considered acceptable as long as the collective contribution of all features, together with the intercept, leads to an accurate prediction.%

\begin{figure}
	\centering\includegraphics[]{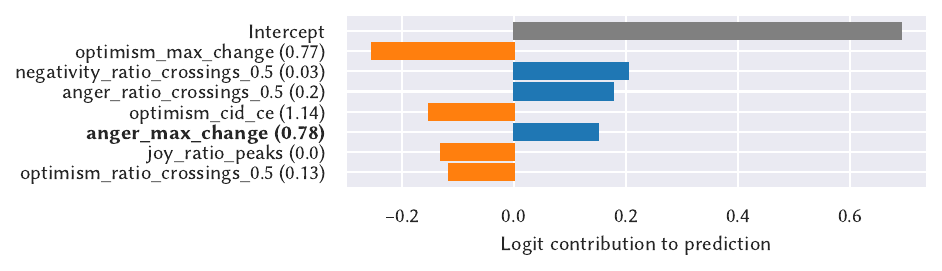}
	\caption{The logit contribution of the most contributing features in the incongruity theory classifier to test joke 194: \textit{If I was a vampire hunter, I'd kill the vampires by inviting them over to my house and serving garlic bread. No one can resist that stuff.} Orange contributions are negative, whereas blue contributions are positive. The intercept is the logit of the prediction that the model will make when all the features take their average values. The feature `maximal change in anger' is formatted in bold, referring to a specific evaluation of the feature function in Figure \ref{fig:featurefunction}.}
	\label{fig:localexample_barplot}
\end{figure}

\begin{figure}
	\centering\includegraphics[]{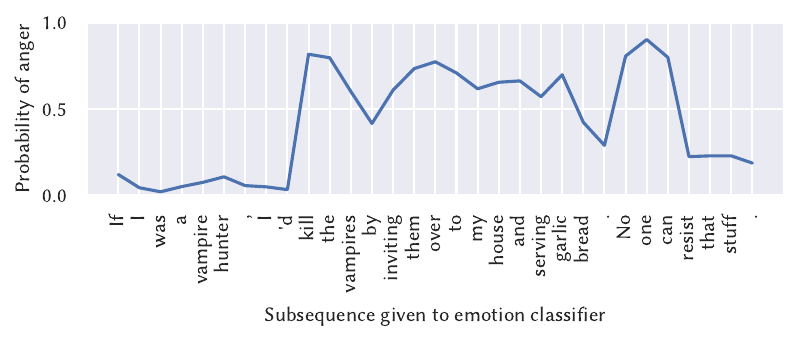}
	\caption{The time series representing the anger probabilities of test joke 194 (Figure \ref{fig:localexample_barplot}), using a subsequence-based approach and the TweetNLP emotion classifier~\cite{camacho-collados-etal-2022-tweetnlp}. The maximal change of two anger probabilities is 0.78.}
	\label{fig:localexample}
\end{figure}

Figure \ref{fig:localexample_barplot} provides a visualization of the seven features that contribute most to the incongruity theory classifier's predictions for a sample test joke. These contributions are evaluations of the feature functions (which are shown on the y-axis), as illustrated by the orange dot in Figure \ref{fig:featurefunction} for the `maximal change in anger' feature. The time series representing the anger probabilities in the joke is visualized in Figure \ref{fig:localexample}. This illustrates the semantic meaning of the incongruity proxy features that have bursts of anger as the underlying hypothesis.

\subsection{EQ5: Identifying the Most Influential Proxy Features to Represent Humor Theory}
Proxy features can be aggregated on a global level in each humor classifier, representing a collective summary of individual local instance results from the training set, and providing an overview of what the humor theory classifier has learned. Highly influential (quantitative) proxy features suggest that if matching with reality (\textbf{EQ3}), their corresponding (qualitative) hypotheses play an important global role in capturing the relevant humor theory for predictive performance in this implementation.
\begin{figure}
	\centering\includegraphics[]{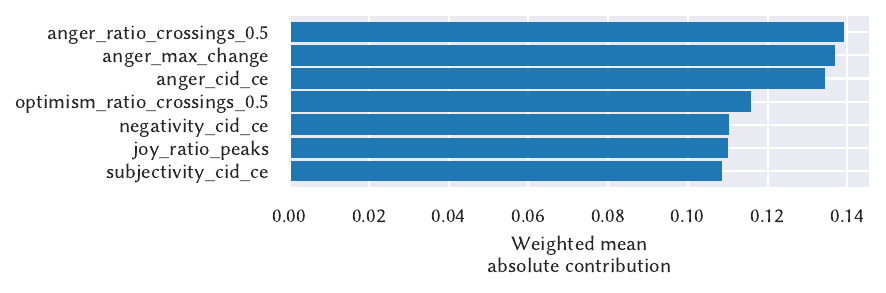}
	\caption{The seven top weighted mean absolute logit contributions in the incongruity theory classifier on the training set.}
	\label{fig:globalinterpretability}
\end{figure}
Figure \ref{fig:globalinterpretability} visualizes the seven proxy features with the greatest weighted absolute average impact on training set classification in the incongruity theory classifier. The top three global features represent the hypothesis that large changes in anger correlate with a higher likelihood of incongruity.

\section{Conclusion}\label{sec:conslusion}
The paper presents the THInC (Theory-driven Humor Interpretation and Classification) framework, a novel humor detection framework that is fully grounded in linguistic humor theories, contributing to bridging a long-standing gap between humor research and computational humor detection. It approaches humor detection as a binary classification problem through an ensemble model of GA$^2$M classifiers. Each classifier embodies a distinct humor theory via a customized feature engineering workflow. This allows for representing humor theories in a straightforward, hypothesized manner. The ensemble then combines their predictive strengths for superior performance.

We created an implementation of the framework within the limitations imposed by today's available humor theories and computational means. Addressing the first research question about the performance of such theory-informed systems (\textbf{RQ1}), our implementation demonstrates that competitive humor detection, with an F1 score of 0.85, is achievable while fully basing all model components on established humor theories (\textbf{EQ1}). In our implementation, the most contributing humor theories to the ensemble results, within the limitations of the proxy features, are the relief theory and the incongruity theory (\textbf{EQ2}).

Addressing the second research question about tracing the learned feature functions to humor theory (\textbf{RQ2}), the framework enables transparent validation of theory-based hypotheses and the efficacy of their numerical derived proxies, laying the first stepping stone towards validating humor theories.
Examining humor theory classifier features provides insights into whether and how well proxy features represent the theoretical concepts (\textbf{EQ3}). The insights also reveal if a new hypothesis or a more effective proxy feature should be sought. Interpretability on local instances shows how specific instances manifest a theory semantically (\textbf{EQ4}). Aggregating local interpretabilities globally allows us to identify the most influential proxies overall in encoding each theory (\textbf{EQ5}).

Our work contributes to bridging computational and theoretical humor research, not only advancing the field of computational humor but also providing a foundation for future explorations into theory-informed recognition of humor.

\section{Future Work}\label{sec:futurework}
While this framework is a first step towards automatic humor detection that integrates humor theories, it contains certain limitations that can and should be addressed in future research. 
One notable limitation is the framework's current inability to fully represent simultaneous scenarios, a critical element in various humor theories. Future research could explore how large language models might be more effectively employed to capture and integrate these multiple scenarios into proxy features.

Additionally, while the current theoretical foundation of our framework is quantitatively verifiable, it is sometimes challenged by the inherent ambiguity present in humor theories.  Efforts to refine these theories into less ambiguous forms could substantially improve the theoretical underpinning and computational implementation, reducing or removing the need to rely on hypotheses.

Another current limitation lies in the handling of semantics and context. Future work could improve this by integrating the latest advancements in language models into the computational tools used for measuring the theories. As a test bed for this, adversarial examples can be leveraged to analyze possible spurious out-of-domain correlations with humor features and make the framework more robust. 

Furthermore, a deeper analysis of how various humor theories perform across different joke genres within the ensemble model could provide valuable insights into how to reconcile divergent or conflicting theories, thereby broadening the framework’s applicability.

Finally, our framework focused on learning and backtracing associations between humor theory features and detection performance, rather than on the causality of humor -- that is, which specific features cause a joke to be perceived as funny. By focusing on causality, future research could advance humor understanding. This approach would pave the way for a more comprehensive and empirically validated computational humor theory.

\section*{Ethics Statement}
We acknowledge that humor is subjective and culture-specific, so some jokes may be offensive to certain people. The interpretability of our framework allows the assessment of potential biases, enabling modifications to improve fairness.

\section*{Acknowledgements}
The authors want to thank Luc De Raedt for supervising the master's thesis from which this project originated, as well as the reviewers for their valuable comments and suggestions on early versions of this paper.
ART and VDM received funding from a starting grant at KU Leuven. VDM received funding from the Flemish Government under the "Onderzoeksprogramma Artificiële Intelligentie (AI) Vlaanderen" programme.
TW received a grant from Internal Funds KU Leuven (PDMT2/23/050) and as a fellow of the Research Foundation-Flanders (FWO-Vlaanderen, 11C7720N).

\bibliography{paper}

\appendix
\newpage
\section{Features used in our implementation}\label{app:features}

\begin{table}[htp]
    \caption{The 25 features used in our implementation of the superiority theory classifier using the THInC framework.}
	\label{tab:features_superiority}
    \begin{tabular}{>{\raggedright\arraybackslash}p{0.22\linewidth}  >{\raggedright\arraybackslash}p{0.22\linewidth} >{\raggedright\arraybackslash}p{0.22\linewidth}  >{\raggedright\arraybackslash}p{0.22\linewidth} }
    	\toprule
        Computational tool & Time series & Manifestation & Derived proxy features \\ \midrule
        TweetNLP offensive language identification \cite{camacho-collados-etal-2022-tweetnlp} & Offense, subsequence-based & Increasing offensive language towards others. & - Linear fit slope\newline- Linear fit standard error\newline- Skewness\newline- Symmetry looking\newline- Aggregated linear trend\newline- Ratio of crossings at 0.9\newline- Ratio of crossings at 0.5 \\ \hline
        Author stance detection model \cite{stancemodel} & Attack, subsequence-based & Increasing aggression or confrontation, enhancing the group cohesion of those who align against the target & - Linear fit slope\newline- Linear fit standard error\newline- Skewness\newline- Symmetry looking\newline- Aggregated linear trend\newline- Ratio of crossings at 0.9\newline- Ratio of crossings at 0.5 \\ \hline
        Hate detection model \cite{vidgen2021learning} & Hate, subsequence-based & Increasing hate speech towards perceived inferiors & - Linear fit slope\newline- Linear fit standard error\newline- Skewness\newline- Symmetry looking\newline- Aggregated linear trend\newline- Ratio of crossings at 0.9\newline- Ratio of crossings at 0.5 \\ \hline
        TweetNLP sentiment analysis \cite{camacho-collados-etal-2022-tweetnlp} & Neutrality, subsequence-based & Absence of neutrality & - Absolute energy\newline- Mean absolute change \\ \hline
        TweetNLP sentiment analysis \cite{camacho-collados-etal-2022-tweetnlp} & Positivity, subsequence-based & No consistent positivity (which would not align with superiority) & - Large standard deviation \\ \hline
        TweetNLP sentiment analysis \cite{camacho-collados-etal-2022-tweetnlp} & Negativity, subsequence-based & No consistent negativity (this would be no humor, just negativity) & - Large standard deviation \\ \bottomrule
    \end{tabular}
\end{table}

\begin{table}
    \caption{The 48 features used in our implementation of the incongruity theory classifier using the THInC framework.}
	\label{tab:features_incongr}
    \begin{tabular}{>{\raggedright\arraybackslash}p{0.22\linewidth}  >{\raggedright\arraybackslash}p{0.22\linewidth} >{\raggedright\arraybackslash}p{0.22\linewidth}  >{\raggedright\arraybackslash}p{0.22\linewidth} }
    	\toprule
        Computational tool & Time series & Manifestation & Derived proxy features \\ \midrule
        Probabilities outputted by Llama 2 \cite{touvron2023llama} & Llama probabilities, token-based & Disruption in normal expectancy shown by sudden changes in probabilities & - Max change\newline- CID CE (complexity estimate)\newline- Ratio of crossings at 0.5\newline- Ratio of wavelet peaks\newline- Ratio of peaks\newline- Ratio beyond 2 sigma \\ \hline
        TweetNLP sentiment analysis \cite{camacho-collados-etal-2022-tweetnlp} & Positivity, subsequence-based & Sudden spikes in positivity in contexts where it is not anticipated & - Max change\newline- CID CE\newline- Ratio of crossings at 0.5\newline- Ratio of wavelet peaks\newline- Ratio of peaks\newline- Ratio beyond 2 sigma \\ \hline
        TweetNLP sentiment analysis \cite{camacho-collados-etal-2022-tweetnlp} & Negativity, subsequence-based & Unexpected drops or increases in negativity that counteract the expected emotional tone & - Max change\newline- CID CE\newline- Ratio of crossings at 0.5\newline- Ratio of wavelet peaks\newline- Ratio of peaks\newline- Ratio beyond 2 sigma \\ \hline
        TweetNLP emotion recognition \cite{camacho-collados-etal-2022-tweetnlp} & Joy, subsequence-based & Bursts of joy that break from the narrative or emotional flow & - Max change\newline- CID CE\newline- Ratio of crossings at 0.5\newline- Ratio of wavelet peaks\newline- Ratio of peaks\newline- Ratio beyond 2 sigma \\ \hline
        TweetNLP emotion recognition \cite{camacho-collados-etal-2022-tweetnlp} & Optimism, subsequence-based & Abrupt transitions to optimism in seemingly unsuitable or unexpected scenarios & - Max change\newline- CID CE\newline- Ratio of crossings at 0.5\newline- Ratio of wavelet peaks\newline- Ratio of peaks\newline- Ratio beyond 2 sigma \\ \hline
        TweetNLP emotion recognition \cite{camacho-collados-etal-2022-tweetnlp} & Sadness, subsequence-based & Placement of sadness in a context that typically calls for happiness & - Max change\newline- CID CE\newline- Ratio of crossings at 0.5\newline- Ratio of wavelet peaks\newline- Ratio of peaks\newline- Ratio beyond 2 sigma \\ \hline
        TweetNLP emotion recognition \cite{camacho-collados-etal-2022-tweetnlp} & Anger, subsequence-based & Unexpected surges of anger that contradict the prevailing mood or setting & - Max change\newline- CID CE\newline- Ratio of crossings at 0.5\newline- Ratio of wavelet peaks\newline- Ratio of peaks\newline- Ratio beyond 2 sigma \\ \hline
        Subjective bias detection model \cite{subjectivity} & Subjectivity, subsequence-based & Shifts in subjectivity that unexpectedly skew perception in otherwise neutral or objective narratives & - Max change\newline- CID CE\newline- Ratio of crossings at 0.5\newline- Ratio of wavelet peaks\newline- Ratio of peaks\newline- Ratio beyond 2 sigma \\ \bottomrule
    \end{tabular}
\end{table}

\begin{table}
    \caption{The 46 features used in our implementation of the relief theory classifier using the THInC framework.}
	\label{tab:features_relief}
    \begin{tabular}{>{\raggedright\arraybackslash}p{0.22\linewidth}  >{\raggedright\arraybackslash}p{0.22\linewidth} >{\raggedright\arraybackslash}p{0.22\linewidth}  >{\raggedright\arraybackslash}p{0.22\linewidth} }
    	\toprule
        Computational tool & Time series & Manifestation & Derived proxy features \\ \midrule
        TweetNLP emotion recognition \cite{camacho-collados-etal-2022-tweetnlp} & Optimism, subsequence-based & Gradual increase in optimism as tension is released & - Linear fit slope\newline- Mean second derivative\newline- Energy ratio by chunks (segments 0 and 3 of 3)\newline- Mass center\newline- Skewness\newline- Symmetry looking \\ \hline
        TweetNLP emotion recognition \cite{camacho-collados-etal-2022-tweetnlp} & Joy, subsequence-based & Gradual increase in joy as tension is released & - Linear fit slope\newline- Mean second derivative\newline- Energy ratio by chunks (segments 0 and 3 of 3)\newline- Mass center\newline- Skewness\newline- Symmetry looking \\ \hline
        TweetNLP emotion recognition \cite{camacho-collados-etal-2022-tweetnlp} & Anger, subsequence-based & Decrease in anger as tension is released & - Linear fit slope\newline- Mean second derivative\newline- Energy ratio by chunks (segments 0 and 3 of 3)\newline- Mass center\newline- Skewness\newline- Symmetry looking \\ \hline
        TweetNLP emotion recognition \cite{camacho-collados-etal-2022-tweetnlp} & Sadness, subsequence-based & Reduction in sadness as tension is released & - Linear fit slope\newline- Mean second derivative\newline- Energy ratio by chunks (segments 0 and 3 of 3)\newline- Mass center\newline- Skewness\newline- Symmetry looking \\ \hline
        Hate detection model \cite{vidgen2021learning} & Hate, subsequence-based & Decrease in hate speech as tension is released & - Linear fit standard error\newline- Aggregate linear trend\newline- Ratio of crossings at 0.5\newline- Skewness\newline- Symmetry looking\newline- First location of maximum/minimum\newline- Energy ratio by chunks (segment 3 of 3)\newline- Mass center \\ \hline
        Adult language detection model \cite{adultlanguage} & Adult language, subsequence-based & Change in the use of adult language as tension is released & - Linear fit slope/standard error\newline- Skewness\newline- Symmetry looking\newline- First location of maximum/minimum\newline- Energy ratio by chunks (segment 0 of 3)\newline- Mass center\newline- Mean second derivative \\ \bottomrule
    \end{tabular}
\end{table}

\begin{table}
    \caption{The 36 features used in our implementation of the surprise disambiguation model classifier using the THInC framework.}
	\label{tab:features_sd}
    \begin{tabular}{>{\raggedright\arraybackslash}p{0.22\linewidth}  >{\raggedright\arraybackslash}p{0.22\linewidth} >{\raggedright\arraybackslash}p{0.22\linewidth}  >{\raggedright\arraybackslash}p{0.22\linewidth} }
    	\toprule
        Computational tool & Time series & Manifestation & Derived proxy features \\ \midrule
        Probabilities outputted by Llama 2 \cite{touvron2023llama} & Llama probabilities, token-based & Unexpected twist during the resolution of a joke & - Energy ratio by chunks (segment 2 of 2)\newline- Mass center\newline- Linear fit standard error\newline- Ratio beyond 1 sigma\newline- Maximum change\newline- Maximum change timing \\ \hline
        TweetNLP offensive language identification \cite{camacho-collados-etal-2022-tweetnlp} & Offense, subsequence-based & Resolution of a joke gives a change in offense & - Linear fit slope\newline- Mean second derivative central\newline- Energy ratio by chunks (segments 0 and 3 of 3)\newline- Mass center\newline- Skewness\newline- Symmetry looking \\ \hline
        Author stance detection model \cite{stancemodel} & Attack, subsequence-based & Resolution of a joke gives a change in attack & - Linear fit slope\newline- Mean second derivative central\newline- Energy ratio by chunks (segments 0 and 3 of 3)\newline- Mass center\newline- Skewness\newline- Symmetry looking \\ \hline
        ConceptNet to retrieve related words + GloVe word embeddings to compute the pairwise distances between these words \cite{speer2017conceptnet,pennington-etal-2014-glove} & Ambiguity, token-based & Ambiguity in jokes is built up and then clarified & - Mass center\newline- Mass 25th percentile\newline- Skewness\newline- Linear fit slope\newline- Linear fit standard error\newline- Aggregated linear trend \\ \hline
        Confidence in the top part-of-speech by the spaCy tagger \cite{Honnibal_spaCy_Industrial-strength_Natural_2020} & Morphosyntactic ambiguity, token-based & Morphosyntactic ambiguity plays with language rules to create and resolve incongruities. & - Mass center\newline- Mass at 25th percentile\newline- Skewness\newline- Linear fit slope\newline- Linear fit standard error\newline- Aggregated linear trend \\ \hline
        Adult language detection model \cite{adultlanguage} & Adult language, subsequence-based & An introduction of adult language can be the resolution of an incongruity. & - First location of maximum\newline- First location of minimum\newline- Mass center\newline- Energy ratio by chunks (segment 2 of 2) \\ \bottomrule
    \end{tabular}
\end{table}

\end{document}